\documentclass[lettersize,journal]{IEEEtran}

\usepackage{amsmath,amsfonts,amssymb}
\usepackage{bm}

\usepackage{graphicx}
\usepackage{booktabs}
\usepackage{array}
\usepackage{multirow}
\usepackage{makecell}
\usepackage{tabularx}

\usepackage[caption=false,font=normalsize,labelfont=sf,textfont=sf]{subfig}

\usepackage{pgfplots}
\usepackage{pgfplotstable}
\pgfplotsset{compat=1.18}

\usepackage{textcomp}
\usepackage{xcolor}
\usepackage{url}
\usepackage{cite}
\usepackage{verbatim}

\usepackage{stfloats}
\usepackage{float}      
\usepackage{placeins}   

\usepackage[ruled,vlined,linesnumbered]{algorithm2e}
\DontPrintSemicolon
\SetKwInput{KwIn}{Input}
\SetKwInput{KwOut}{Output}
\SetKwComment{tcp}{// }{}

\usepackage{hyperref}
\hypersetup{hypertex=true,
colorlinks=true,
linkcolor=blue,
anchorcolor=blue,
citecolor=blue}

\newcommand{\ie}{\textit{i}.\textit{e}.}

\begin{document}

\title{DeTrack: A Benchmark and Altitude-Aware Dual World Model for Drone-embodied Tracking}

\author{Guyue Hu, Haoming Liu, Siyuan Song, Chenglong Li, Feng Chen, and Jin Tang
\thanks{This work was supported in part by the National Natural Science Foundation of China (No. 62506003, No. 62376004), the Anhui Provincial Natural Science Foundation (No. 2408085QF201), the Natural Science Research Major Project of Anhui Provincial Education Department (2025AHGXZK20020), and the Open Project of Anhui Provincial Key Laboratory of Intelligent Detection and Diagnosis for Traffic Infrastructure (No. KY-2025-03). (Corresponding author: Chenglong Li)}
\thanks{Guyue Hu, Haoming Liu, Siyuan Song and Chenglong Li are with the School of Artificial Intelligence, State Key Laboratory of Opto-Electronic Information Acquisition and Protection Technology, and Anhui Provincial Key Laboratory of Security Artificial Intelligence, Anhui University, Hefei 230601, China (e-mail: guyue.hu@ahu.edu.cn, lhm@stu.ahu.edu.cn, ssy136@stu.ahu.edu.cn, lcl1314@foxmail.com).}
\thanks{Feng Chen is with the Hefei Si Valley Technology Development Co., Ltd, and also with the Institute of Embodied Intelligence, Anhui University, Hefei 230601, China. (e-mail: fengchen@ahu.edu.cn).}
\thanks{Jin Tang is with the School of Computer Science and Technology, State Key Laboratory of Opto-Electronic Information Acquisition and Protection Technology, and Anhui Provincial Key Laboratory of Multimodal Cognitive Computation, Anhui University, Hefei 230601, China (e-mail: tangjin@ahu.edu.cn).}
}

\maketitle

\begin{abstract}
Aerial object tracking has wide applications in public safety, emergency rescue, wildlife monitoring, etc. However, existing aerial object tracking is mainly limited to the passive tracking paradigm with 2D video sequences pre-recorded in fixed camera locations or along pre-defined flight paths, lacking drone-embodied scene perception, egocentric interaction, and movement control in dynamic 3D scenes. In this paper, we define a novel drone-embodied tracking (DeTrack) task and build a large-scale benchmark with evaluation metrics for this task. It refers to tracking targets in interactive 3D scenes with online drone-egocentric observation and active control in a closed loop. The DeTrack benchmark differs from conventional passive aerial tracking in five characteristics: (1) drone-embodied closed-loop tracking, (2) target tracking coupled with obstacle avoidance, (3) spatio-temporally composite occlusion in 3D scenes, (4) altitude-mediated contradiction between visibility and safety, (5) extensive and extreme scale variation. Besides, the drone-embodied tracking faces intrinsic altitude-mediated contradiction between visibility and safety: the view field is wide while the target detail is weak at high flight altitude, and the target detail is sufficient while the obstacles are dense at low flight altitude. Therefore, we further propose a novel altitude-aware dual worlds (AaDWorlds) for drone-embodied tracking. It consists of an altitude-aware perception (AaP) module and dual world models (DWM). Specifically, the altitude-aware perception module adaptively encodes, couples, and decodes altitude-aware representations. The dual world models imagine future states under both high- and low-altitude regimes. Eventually, the imagined future states (from DWM) and pseudo egocentric observations (from AaP) at both high and low altitudes effectively complement the original drone-egocentric observation, simultaneously facilitating the view field, target detail, and obstacle avoidance during drone-embodied tracking.
\end{abstract}


\section{Introduction}
\label{sec:intro}
Object tracking has wide applications in the fields of emergency rescue, traffic inspection, ecological protection, anti-drone confrontation, and so on~\cite{wu2021uav_survey,azar2021drone_drl,chan2024dronesim_survey,liu2022monkeytrail}. With the rapid growth of low-altitude economy and unmanned aerial vehicles (UAVs), aerial object tracking has achieved significant progress. Representative aerial object tracking benchmarks, such as UAV123~\cite{mueller2016uav123}, UAVDT~\cite{du2018uavdt}, and VisDrone~\cite{zhu2018visdrone,zhu2018visdrone_vdt}, have promoted standardized evaluation on aerial videos. The classical aerial object tracking approach has evolved from correlation-filter~\cite{henriques2015kcf}, Siamese-based trackers~\cite{bertinetto2016siamfc,li2019siamrpn++}, and more recent Transformer-based trackers~\cite{chen2021transt,yan2021stark,ye2022ostrack}. 

\begin{figure}[t]
\centering
\vspace{-4pt}
\captionsetup{font=footnotesize, skip=2pt}

\includegraphics[
  width=\linewidth,
  trim=4pt 2pt 4pt 2pt,
  clip
]{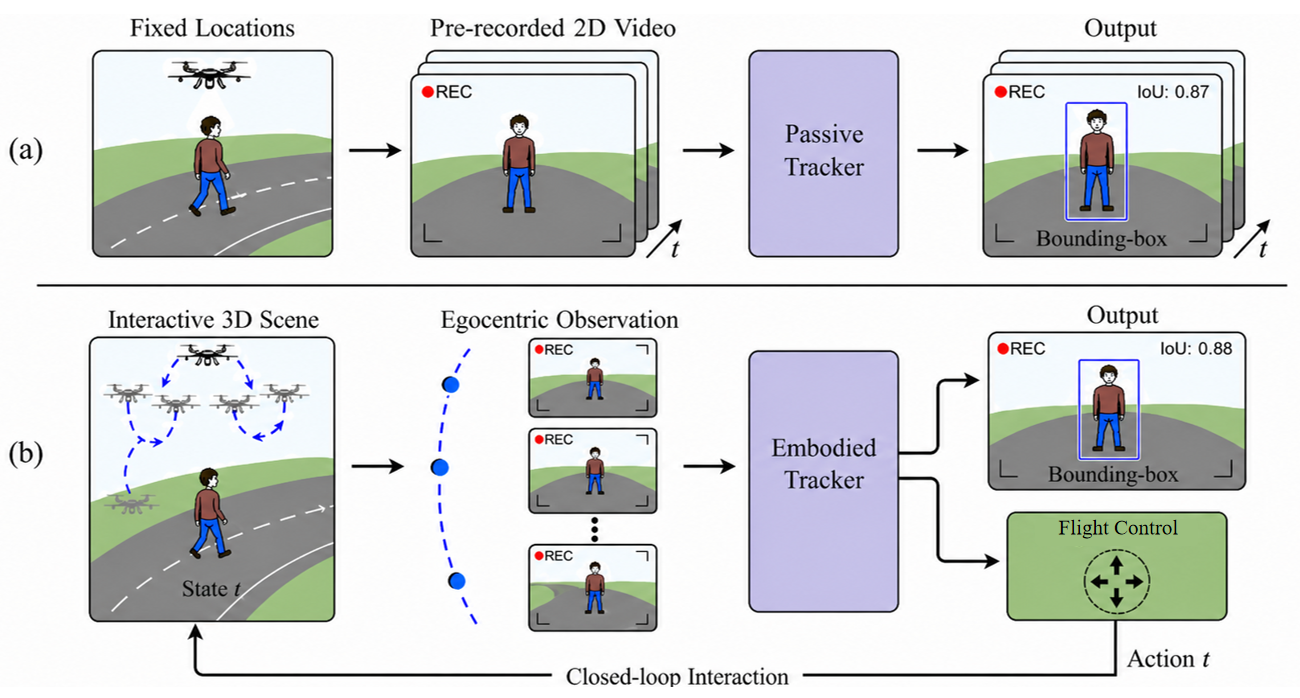}
\vspace{-3pt}
\caption{Paradigm comparison between passive aerial tracking and drone-embodied tracking. (a) In the passive tracking paradigm, trackers process pre-recorded aerial videos captured from fixed camera locations or pre-defined flight paths. (b) In the drone-embodied tracking paradigm, the drone actively interacts with an interactive 3D scene and uses online egocentric observations to track the target and update its flight action in a closed loop.}
\label{fig:comparison}
\vspace{-8pt}
\end{figure}

\textit{However, existing works and benchmarks for aerial object tracking are mainly confined to the passive tracking paradigm} (Fig.~\ref{fig:comparison}a), where object trackers are evaluated on 2D aerial video sequences recorded from fixed locations or along pre-defined flight paths. As a result, drones are merely treated as passive cameras without active scene perception, motion control, and environment interaction. However, the real-world operating environment for drones is typically wide, dynamic, and complicated (obstacle-dense, occlusion-frequent), which requires active abilities of environment perception and movement control and urgently calls for a new drone-embodied tracking paradigm (Fig.~\ref{fig:comparison}b).

\begin{figure*}[tpb]
  \centering
  \includegraphics[width=1\textwidth]{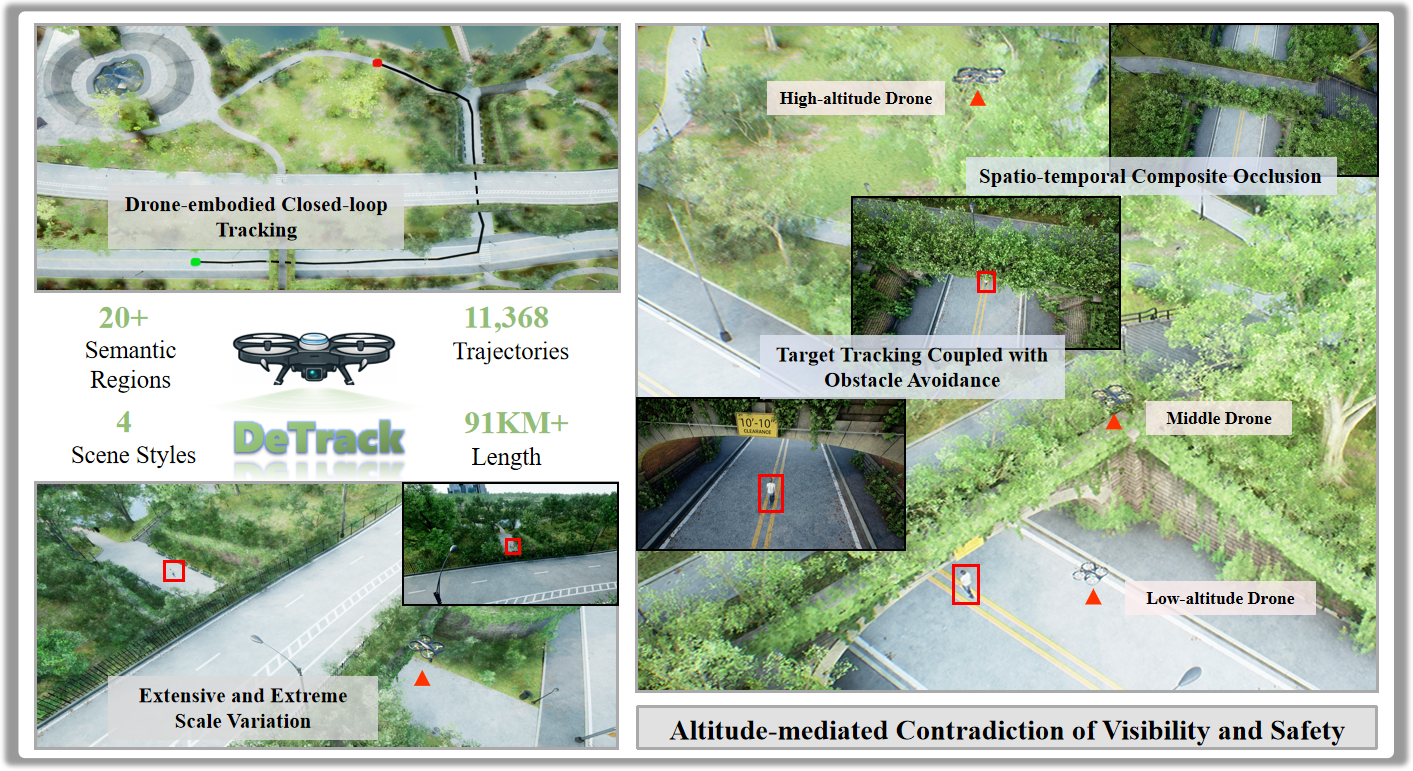}
  \vspace{-6pt}
  \caption{Overview statistics and typical challenges of the proposed drone-embodied tracking (DeTrack) benchmark.}
  \label{fig:challenges}
  \vspace{-14pt}
\end{figure*}

To activate and advance the research and application of real-world drone object tracking, we define a novel drone-embodied tracking task and build the first large-scale benchmark with evaluation metrics for this task. The DeTrack benchmark is built in an interactive 3D engine and contains 11,368 target trajectories with an average length of 80.09 m for tracking. The drone-embodied tracking task requires the embodied drone to track these targets in interactive 3D scenes with online egocentric observation and active movement control. As illustrated in Fig.~\ref{fig:challenges},  the DeTrack task and benchmark differ from conventional passive aerial tracking by five characteristics. 
(1) \textit{drone-embodied closed-loop tracking}, where historical actions shape future observations and the observations shape the following action in turn;
(2) \textit{target tracking coupled with obstacle avoidance}, which requires simultaneous target following and safe navigation;
(3) \textit{spatio-temporally composite occlusion in 3D scenes}, arising from both static scene structures and dynamic distractors;
(4) \textit{altitude-mediated contradiction between visibility and safety}, which couples target observability with obstacle interactions;
(5) \textit{extensive and extreme scale variation}, caused by continuous altitude changes during flight. Taken together, these task characteristics make DeTrack a challenging task and benchmark in interactive 3D environments.

\textit{Besides, the 3D drone-embodied tracking task faces an intrinsic altitude-mediated contradiction of visibility and safety}. Specifically, at high flight altitudes, the drone enjoys a wider view and fewer obstacle interactions, but the target becomes small and visually ambiguous. At low flight altitudes, the target appears with richer details and better visibility, but the drone faces denser obstacles, more frequent occlusions, and higher collision risk. To alleviate this intrinsic contradiction, we further propose a novel altitude-aware dual worlds framework for drone-embodied tracking (AaDWorlds), which consists of an altitude-aware perception module (AaP) and dual world models (DWM). Specifically, the AaP is designed to encode, couple, and decode altitude-aware observation representations, while the DWM imagines future states in both high- and low-altitude regimes. During inference, with the aid of imagined future states from DWM and pseudo egocentric observations from AaP at both high and low altitudes, our AaDWorlds simultaneously facilitates the view field, target detail, and obstacle avoidance, effectively alleviating the intrinsic altitude-mediated contradiction between visibility and safety during drone-embodied tracking.

In summary, the main contributions of this paper are 4-folds. (1) We define a novel drone-embodied tracking task that tracks targets in interactive 3D scenes with online egocentric observation and active movement control, which differs from conventional passive aerial tracking in five characteristics. (2) We build the first large-scale benchmark and corresponding evaluation metrics for drone-embodied tracking, termed DeTrack, which provides an initial benchmark and evaluation protocols to activate and advance this task. (3) We propose a novel altitude-aware dual worlds approach for the drone-embodied tracking task, which effectively tackles the intrinsic altitude-mediated contradiction between visibility and safety in closed-loop drone-embodied tracking. (4) The proposed AaDWorlds approach achieves promising performance on the DeTrack benchmark across all evaluation metrics, demonstrating the effectiveness and superiority of our approach.

\section{Related Works}
\label{sec:related}

\subsection{Passive Drone Tracking}
Most existing drone object tracking studies are based on passive and frame-based benchmarks. Representative datasets such as UAV123~\cite{mueller2016uav123}, UAVDT~\cite{du2018uavdt}, and VisDrone~\cite{zhu2018visdrone,zhu2020drones_challenge} provide pre-recorded aerial videos with ground-truth bounding boxes for offline evaluation. These benchmarks cover diverse scenes, target motions, and imaging conditions, and have greatly promoted the development of drone tracking. Recent datasets further improve scene scale and realism, such as city-scale or long-term tracking settings~\cite{zhu2020drones_challenge,zheng2023uav2uav,wu2021uav_survey,zhao2022antiuav}. However, they still mainly evaluate passive tracking, where the camera trajectory is fixed and the drone does not actively interact with the environment. On top of these benchmarks, drone tracking methods have evolved from correlation-filter and Siamese-based trackers~\cite{henriques2015kcf,bertinetto2016siamfc} to stronger deep models such as SiamRPN++~\cite{li2019siamrpn++} and Transformer-based trackers including TransT~\cite{chen2021transt}, STARK~\cite{yan2021stark}, and OSTrack~\cite{ye2022ostrack}. These methods achieve strong per-frame localization accuracy and robustness under challenges such as motion blur, appearance variation, and background distractors. Beyond RGB-only tracking, a recent RGBT tracking study further addresses modality-missing scenarios through missingness-aware prompting~\cite{hu2025missingness}. Nevertheless, these methods are still mainly designed for passive visual tracking on prerecorded videos, and do not explicitly model how drone motion, altitude change, or viewpoint adjustment will influence future target visibility and tracking performance.

\subsection{Embodied Perception}
Embodied perception studies how an agent actively adjusts its motion or viewpoint to improve downstream perception and decision-making. In object tracking, prior works have explored active tracking with learned camera control, where deep reinforcement learning maps egocentric observations to camera motions for continuous target following~\cite{luo2018end2end_active_tracking}. Subsequent studies extend this idea to 3D scenes, multiple targets, and aerial platforms, applying DRL to air-to-ground tracking, drone navigation, and vision-based target following from onboard images~\cite{nguyen2023ldvrl,huang2019drlnavigation,azar2021drone_drl,feng2024uav_searchtrack}. These works show the potential of embodied control for tracking, but most of them still rely on short-horizon, model-free policies and do not explicitly reason about how altitude changes affect future visibility, occlusion, and collision risk. Meanwhile, embodied AI platforms and active perception research provide important support for this line of work. High-fidelity simulators such as Habitat and related platforms~\cite{savva2019habitat,kolve2017ai2thor,xia2018gibson}, as well as UE- and AirSim-based drone simulators~\cite{shah2017airsim,krishnan2019airlearning,chan2024dronesim_survey}, enable reproducible studies of navigation, obstacle avoidance, and vision-based control in interactive 3D environments. Beyond tracking, viewpoint planning and next-best-view methods investigate how to move sensors to obtain more informative observations~\cite{bircher2016receding,galceran2013cppsurvey}. However, existing embodied benchmarks mainly focus on ground navigation, exploration, or static-view planning, and seldom provide a standardized drone-specific tracking setting empowered with unified closed-loop evaluation. 

\section{Drone-embodied Tracking}
\subsection{Task Formulation}
Conventional passive aerial object tracking tracks the target in pre-recorded 2D video sequences at fixed camera locations or along pre-defined flight paths. In contrast, the proposed drone-embodied tracking task (DeTrack) tracks targets in large interactive 3D scenes with online drone-egocentric observations $(o_{t-K+1:t})$ and active flight control in a closed loop. Specifically, assume that a target moves along a trajectory $(\tau)$ in an interactive 3D scene. The drone-embodied agent exploits the initial target template $(x^{\mathrm{tmp}})$ and online egocentric observations $(o_{t-K+1:t})$ to continuously control the drone, follow the target, and predict the target bounding box $(B_t^{\mathrm{trk}})$. It is a challenging task that involves scene perception, flight control, passive tracking, etc.

\subsection{DeTrack Benchmark} 
\begin{figure}[h]
    \centering
    \includegraphics[
        width=0.98\linewidth,
        trim=0 6 0 6,
        clip
    ]{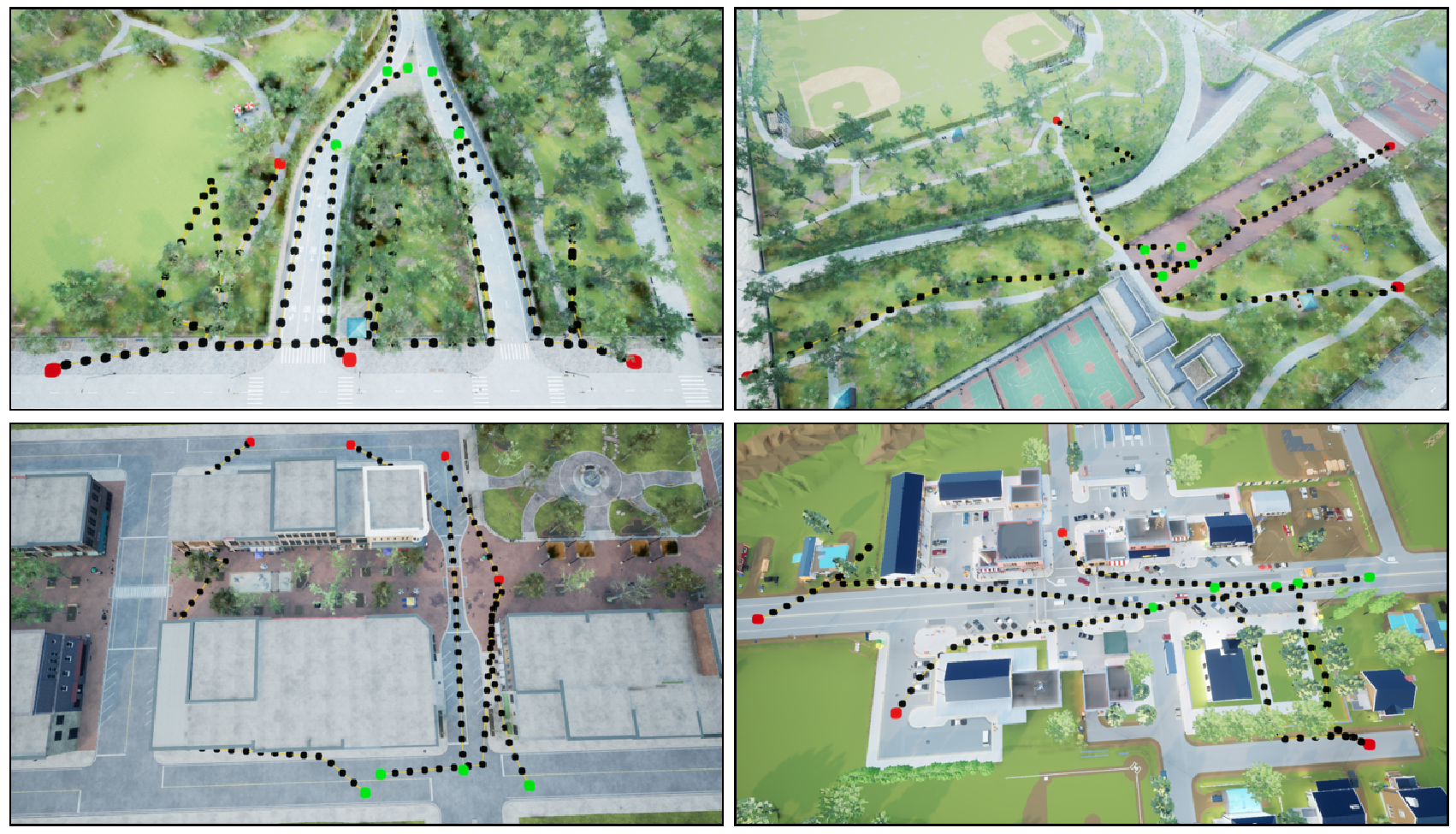}
    \vspace{-6pt}
    \caption{Sample illustration of target trajectories in the DeTrack benchmark.}
    \label{fig:target_trajectory_simulator}
    \vspace{-8pt}
\end{figure}

The DeTrack benchmark for drone-embodied tracking is built in an interactive Unreal Engine 4.27 simulator~\cite{epic2021ue427} with the AirSim plugin~\cite{airsim_docs,airsim2018fsr}. Each sample for drone-embodied target tracking is a predefined moving trajectory in the form of way-point sequences for a target (such as the trajectory samples illustrated in Fig.~\ref{fig:target_trajectory_simulator}). In total, our trajectory library contains 11,368 trajectories. Each trajectory consists of 2 to 99 waypoints, with a mean of 13.84 and a median of 8. In terms of spatial extent, trajectory lengths range from 8.27\,m to 482.74\,m, with a mean length of 80.09\,m and a median length of 69.33\,m (see Fig.~\ref{fig:traj_basic_stats} and Fig.~\ref{fig:traj_geometry_stats}(b) for details). These statistics show that DeTrack covers diverse motion scales and tracking horizons, making it suitable for evaluating embodied tracking under both short-range maneuvers and longer-horizon target following.

The benchmark contains 4 large and representative scenes (Town, Rural, Park, and City), covering scene-level, region-level, domain-level, and target-level diversity (See Fig.~\ref{fig:scene_rendering_conditions} and Fig.~\ref{fig:semantic_region_diversity}). In detail, the scenes differ in global layout and visual statistics, such as building density, vegetation coverage, road topology, and background clutter. Each scene further contains multiple semantic regions with substantial within-scene variation in geometry, occlusion patterns, and background appearance, such as bridges and underpasses, open plazas, dense trees, waterfronts, sports facilities, intersections, and narrow corridors. Each scene is rendered under multiple domain conditions (such as Day, Night, Fog, Sun, Rain, Snow, etc.). Each environment contains primary targets together with additional moving distractors in diverse categories (such as people, vehicles, bicycles, etc.).

\subsection{DeTrack Challenges}
The DeTrack benchmark is tailored for closed-loop drone-embodied tracking in dynamic 3D environments. In contrast to conventional passive aerial tracking, it introduces several task and data challenges (as illustrated in Fig.~\ref{fig:challenges}). (1) Drone-embodied closed-loop tracking: unlike passive object tracking benchmarks, the observations and viewpoints (locations, angles, etc.) of the drone-embodied agent are non-stationary. Its historical actions shape its future egocentric observations, making long-horizon following and target re-acquisition much more challenging.
(2) Target tracking coupled with obstacle avoidance: the drone-embodied agent must simultaneously coordinate target tracking and obstacle avoidance, i.e., keep the target observable and followed while avoiding collisions with surrounding obstacles for safety constraints.
(3) Spatio-temporally composite occlusion: complicated occlusions are composed of static elements (such as buildings, vegetation, and traffic infrastructure) and dynamic entities (vehicles and pedestrians). These occlusions exhibit a mixed nature in spatial visibility and temporal duration, ranging from transient partial obstructions to abrupt and complete disappearance, as well as sustained visibility interruptions over extended durations.
(4) Altitude-mediated contradiction of visibility and safety: the drone flight altitude mediates the difficulty of target perception and obstacle avoidance, inducing an intrinsic contradiction between visibility and safety. In detail, flying at high altitude provides wider views and fewer obstacles, but weakens target visibility and details. In contrast, flying at low altitude provides richer target details and better visibility, but faces more occlusion and obstacles.
(5) Extensive and extreme scale variation: the target sizes in egocentric observations change with the drone altitude substantially and continuously. Low-altitude observations provide relatively large targets, while high-altitude observations even shrink the target to only a few pixels.

\subsection{Evaluation Metrics}
\label{metrics}
To evaluate drone-embodied tracking under closed-loop interaction, we introduce four metrics: visible rate (VR), mean IoU (mIoU), tracking rate (TR), and trajectory success rate (SR), as shown in Table~\ref{tab:metrics}. For a trajectory of length $T$, let $B_t^{\mathrm{gt}}$ and $B_t^{\mathrm{trk}}$ denote the ground-truth and tracker-predicted bounding boxes at step $t$, respectively. Here, $p_t^{\mathrm{vis}}$ is the number of visible target pixels in the segmentation mask, $\tau_{\mathrm{vis}}$ is the visibility threshold, and $\tau_{\mathrm{trk}}$ is the IoU threshold for effective tracking. 
$N_{\mathrm{traj}}$ denotes the number of evaluated trajectories, $T_j$ is the length of the $j$-th trajectory, $d_{xy,T_j}^{(j)}$ is the final horizontal distance between the UAV and the target, and $\tau_d$ is the trajectory success threshold.

\begin{table}[htbp]
\centering
\caption{Evaluation metrics for the drone-embodied tracking task.}
\label{tab:metrics}

\footnotesize
\setlength{\tabcolsep}{4pt}
\renewcommand{\arraystretch}{1.35}

\begin{tabularx}{\linewidth}{@{}>{\bfseries\centering\arraybackslash}p{0.18\linewidth} >{\centering\arraybackslash}X@{}}
\toprule
Metric & Definition \\
\midrule

VR &
\(\textstyle
\frac{1}{T}\sum\nolimits_{t=1}^{T}
\mathbb{I}\!\left(p_t^{\mathrm{vis}}\ge\tau_{\mathrm{vis}}\right)
\) \\

mIoU &
\(\textstyle
\frac{1}{T}\sum\nolimits_{t=1}^{T}
\mathrm{IoU}_t
\) \\

TR &
\(\textstyle
\frac{1}{T}\sum\nolimits_{t=1}^{T}
\mathbb{I}\!\left(p_t^{\mathrm{vis}}\ge\tau_{\mathrm{vis}},\ 
\mathrm{IoU}_t>\tau_{\mathrm{trk}}\right)
\) \\

SR &
\(\textstyle
\frac{1}{N_{\mathrm{traj}}}\sum\nolimits_{j=1}^{N_{\mathrm{traj}}}
\mathbb{I}\!\left(d_{xy,T_j}^{(j)}\le\tau_d\right)
\) \\

\bottomrule
\end{tabularx}
\end{table}

\subsection{Benchmark Details}
\label{sec:benchmark_construction}
Following the benchmark overview above, this subsection provides detailed statistical analysis from two aspects: diverse interactive environments and detailed target trajectory statistics. The interactive environments provide diverse 3D scenes, semantic regions, rendering conditions, and moving distractors for closed-loop drone-embodied tracking. The trajectory statistics summarize the data characteristics of target motions, including trajectory scale, waypoint density, and spatial distribution.

\begin{figure}[tbp]
    \centering
    \includegraphics[
        width=\linewidth,
        trim=0 0 0 0,
        clip
    ]{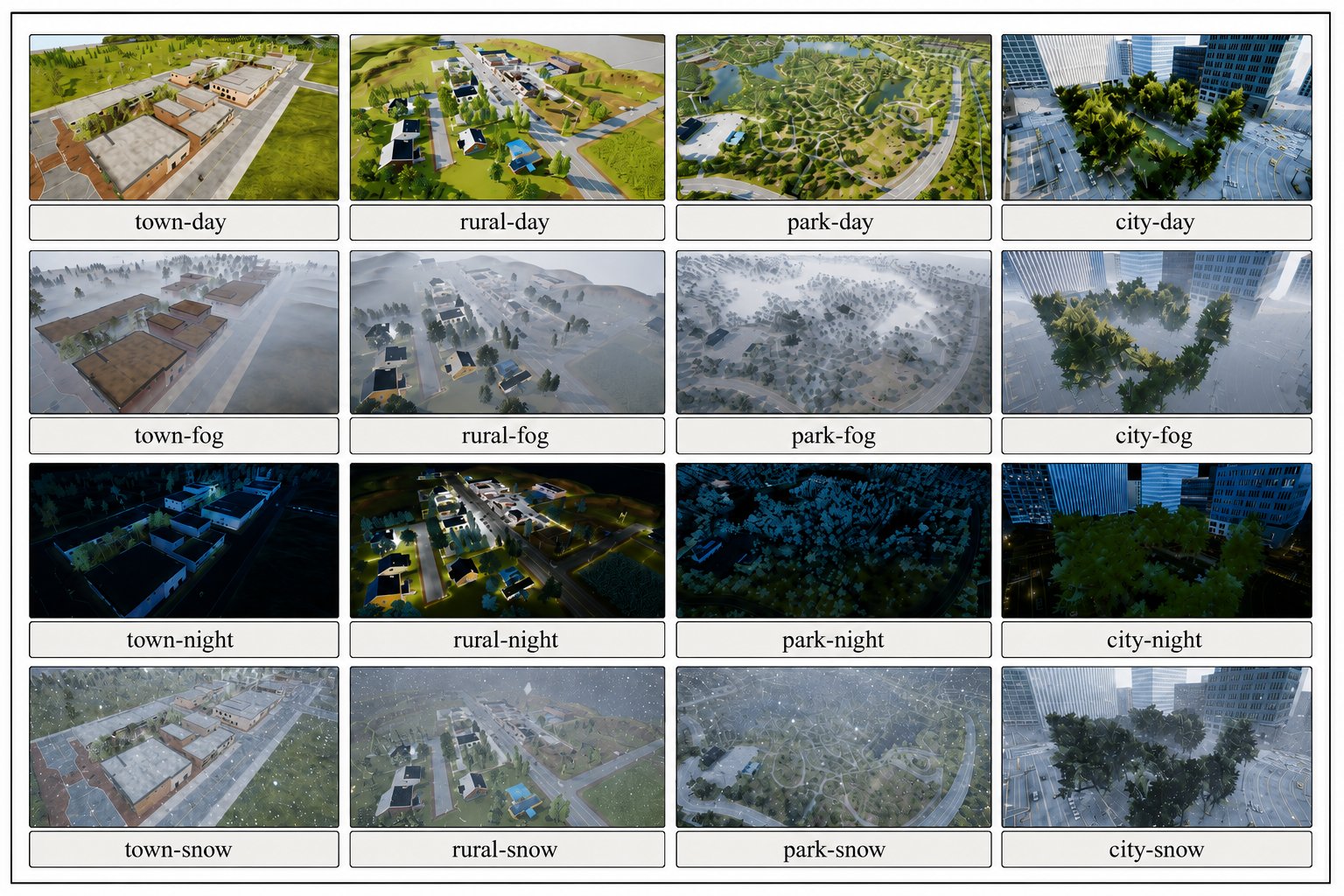}
    \vspace{-4pt}
    \caption{Illustration of scene-style and rendering-condition diversity in the DeTrack benchmark. The figure shows four representative scene styles, including Town, Rural, Park, and City, under different rendering conditions, including day, fog, night, and snow, demonstrating the visual diversity introduced by scene layouts, illumination changes, and weather variations.}
    \label{fig:scene_rendering_conditions}
    \vspace{-8pt}
\end{figure}

(1) \textit{Diverse interactive environments and rendering conditions.}
\phantomsection
Our benchmark environments are constructed in an interactive Unreal Engine 4.27 simulator with the AirSim plugin, covering four representative scene styles: Town, Rural, Park, and City.
These scene styles differ in global layout and visual statistics, such as building density, vegetation coverage, road topology, and background clutter, providing diverse geometric structures and natural occlusion sources across environments.
To further increase visual diversity, each scene is rendered under multiple domain conditions, including different illumination and weather settings.
As shown in Fig.~\ref{fig:scene_rendering_conditions}, the same scene style can exhibit substantially different appearances under day, fog, night, and snow conditions, which introduces additional challenges such as low illumination, reduced visibility, and background variation.

\begin{figure}[h]
    \centering
    \includegraphics[
        width=0.95\linewidth,
        trim=0 0 0 0,
        clip
    ]{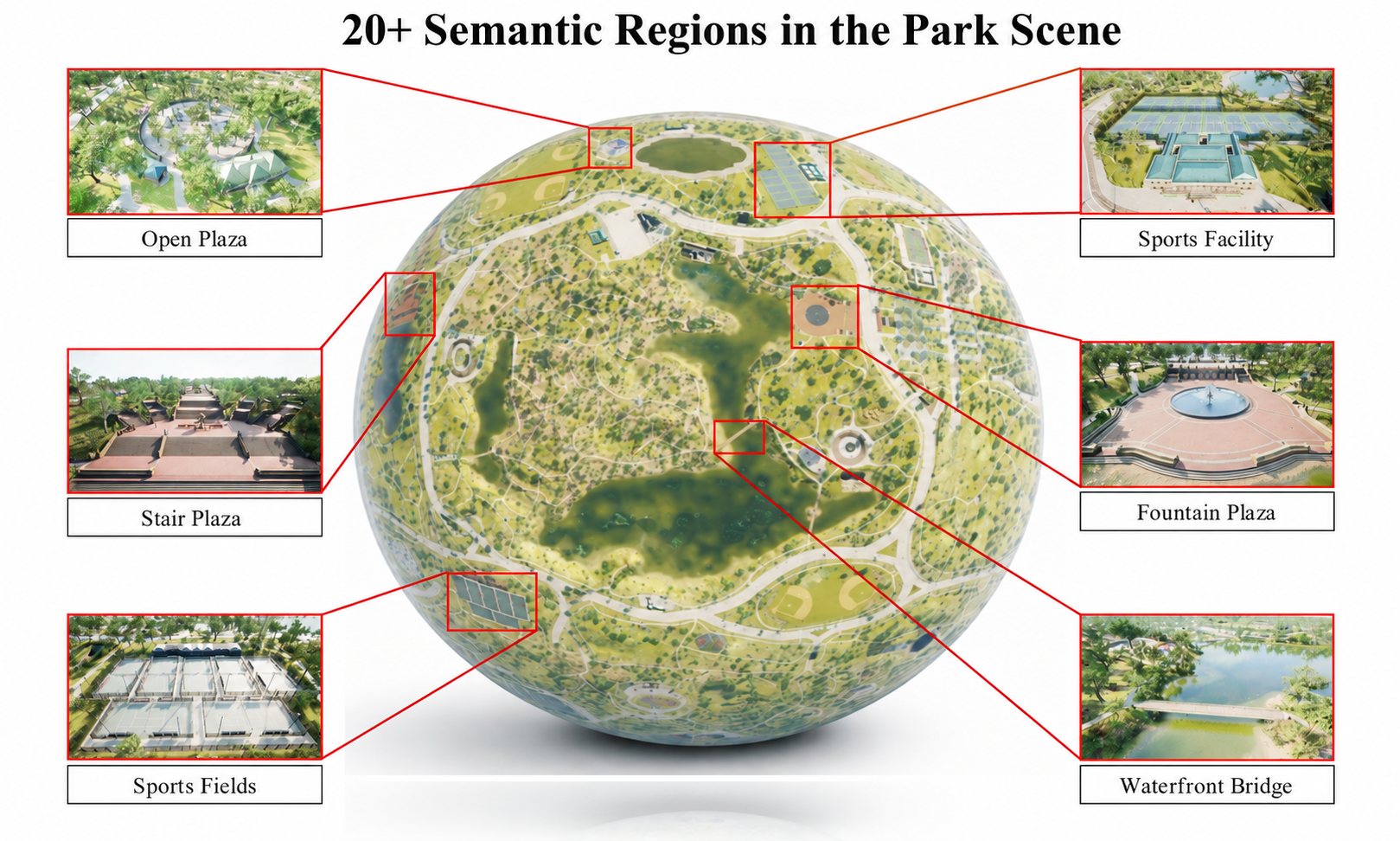}
    \vspace{-6pt}
    \caption{Examples of semantic region diversity within the Park scene. The central map and surrounding thumbnails show representative semantic regions, including open plaza, stair plaza, sports field, sports facility, fountain plaza, and waterfront bridge.
These regions introduce diverse local layouts, occlusion patterns, and background appearances for drone-embodied tracking.}
    \label{fig:semantic_region_diversity}
    \vspace{-8pt}
\end{figure}

In addition to scene-level and rendering-level diversity, each scene contains multiple semantic regions, such as bridges and underpasses, open plazas, dense vegetation, waterfronts, sports fields, intersections, and narrow corridors.
These regions introduce substantial variation in local geometry, occlusion patterns, and background appearance.
Fig.~\ref{fig:semantic_region_diversity} provides representative examples of within-scene semantic region diversity. Meanwhile, the environments contain primary tracking targets together with additional moving distractors, such as pedestrians, vehicles, and bicycles.
This setting makes the benchmark closer to interactive drone operation, where the tracking agent needs to maintain the target while facing visual distractors and dynamic occlusions.

\FloatBarrier

\begin{figure}[tbp]
    \centering
    \includegraphics[
        width=0.98\linewidth,
        trim=0 0 0 0,
        clip
    ]{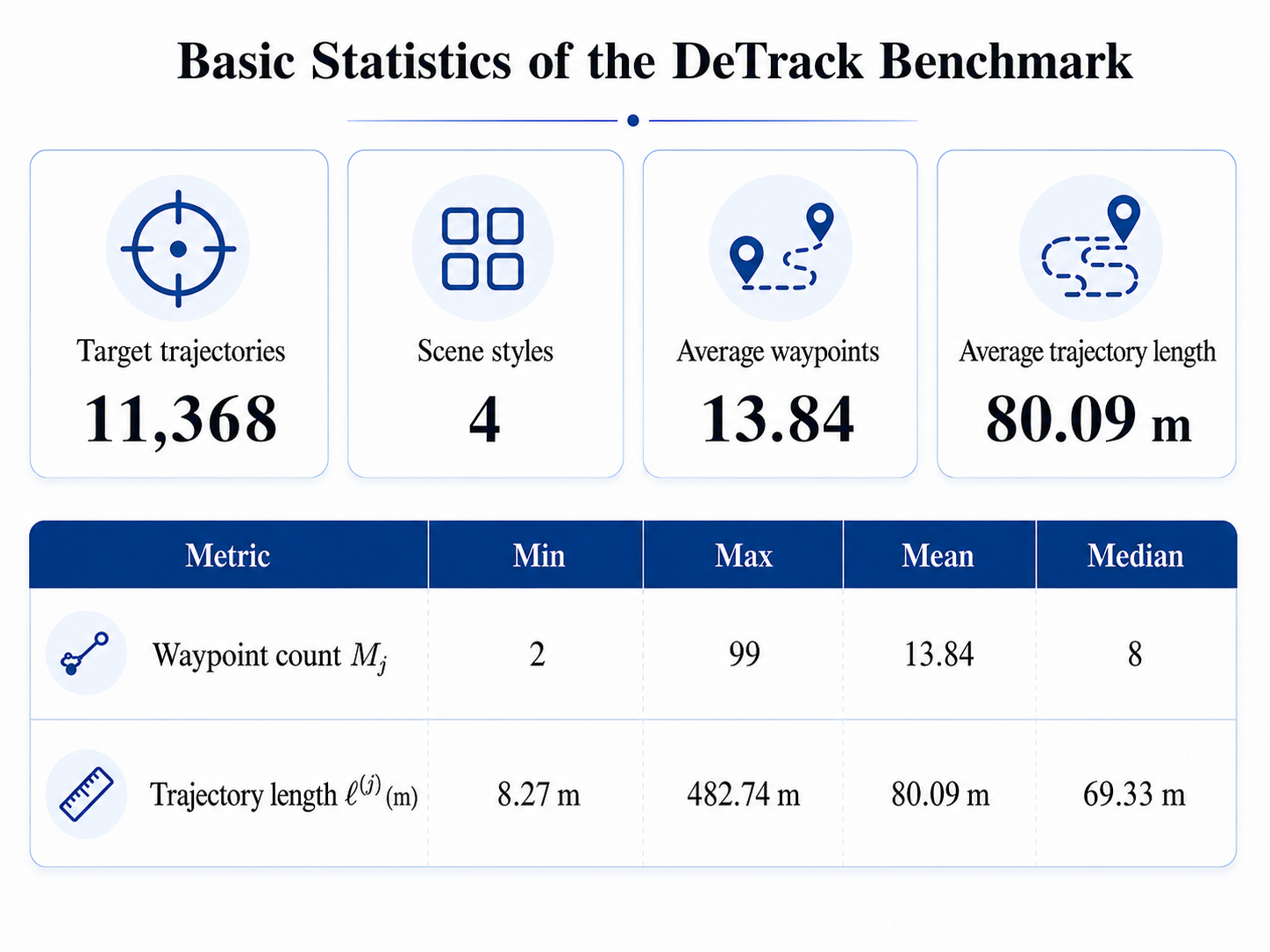}
    \vspace{-6pt}
    \caption{Basic statistics of target trajectory library in the DeTrack benchmark.}
    \label{fig:traj_basic_stats}
    \vspace{-8pt}
\end{figure}

(2) \textit{Detailed target trajectory statistics and analysis}.
\phantomsection
We summarize the basic numerical statistics of the target trajectory library in Fig.~\ref{fig:traj_basic_stats}. These statistics show that the trajectory library covers diverse waypoint densities and motion scales, making it suitable for closed-loop drone-embodied tracking evaluation under different target-following horizons.

\begin{figure}[tbp]
    \centering
    \includegraphics[
        width=0.98\linewidth,
        trim=0 0 0 0,
        clip
    ]{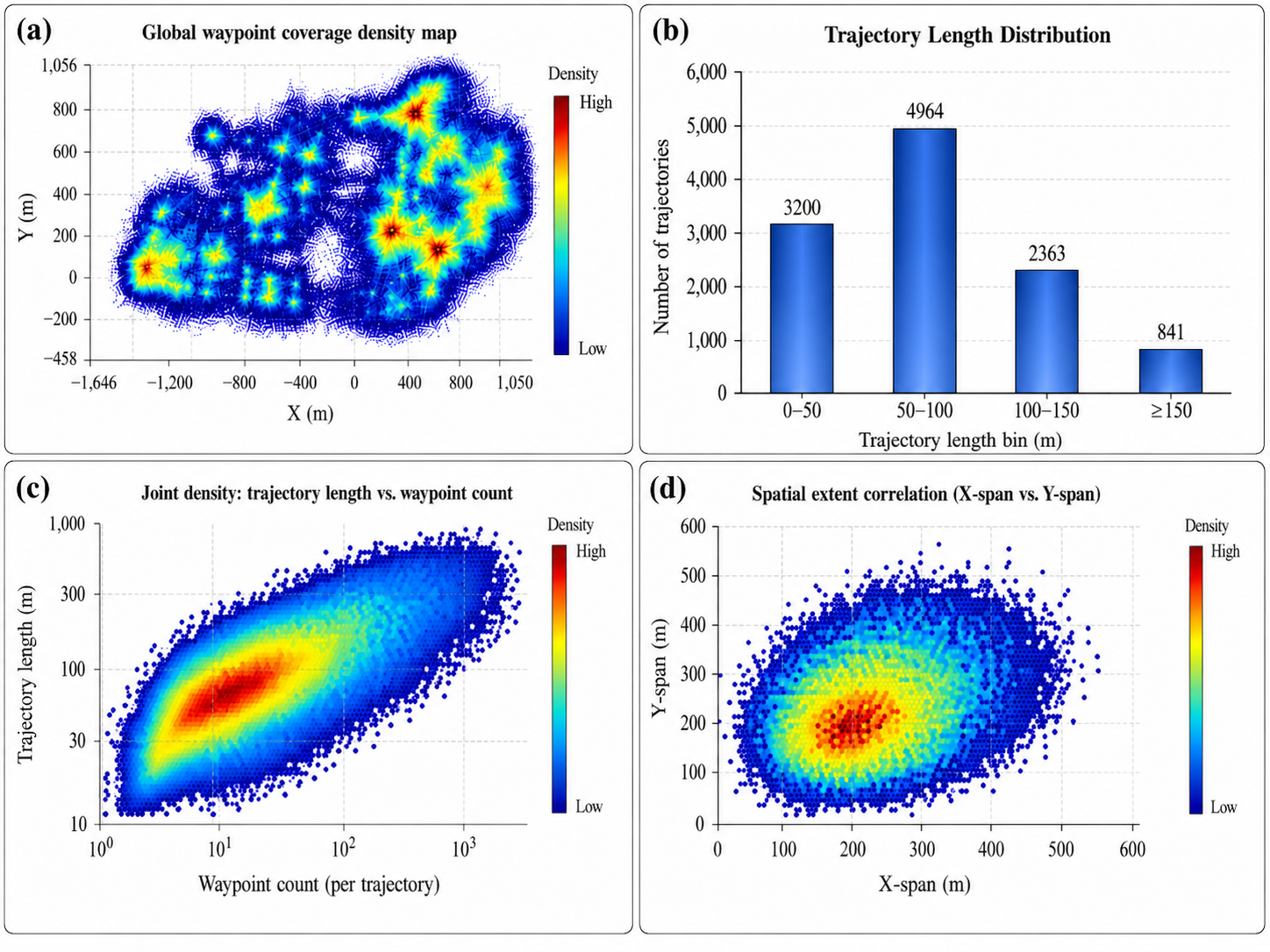}
    \vspace{-6pt}
    \caption{Geometry-level spatial statistics of the DeTrack trajectory library.
    (a) Global waypoint density map showing the spatial coverage of all trajectories.
    (b) Binned distribution of trajectory lengths.
    (c) Joint density between trajectory length and waypoint count.
    (d) Planar spatial extent correlation measured by the \(X\)-span and \(Y\)-span of each trajectory.}
    \label{fig:traj_geometry_stats}
    \vspace{-8pt}
\end{figure}

Beyond the basic numerical statistics in Fig.~\ref{fig:traj_basic_stats}, we further analyze the trajectory library from four geometry-level aspects, as shown in Fig.~\ref{fig:traj_geometry_stats}, including global waypoint coverage, trajectory length distribution, the joint distribution between trajectory length and waypoint count, and planar spatial extent correlation.
The global waypoint coverage indicates that the target trajectories span a broad simulator space.
The trajectory length distribution further reflects the variation of motion scales in the library.
The joint distribution between trajectory length and waypoint count shows the correlation between motion range and waypoint density.
For planar spatial extent correlation, we compute the \(X\)-span and \(Y\)-span of each target trajectory by subtracting the minimum waypoint coordinate from the maximum waypoint coordinate along the corresponding axis.
This statistic further reveals diverse movement ranges on the simulator ground plane.
Taken together, these statistics show that the trajectory library covers diverse motion scales, waypoint densities, and planar movement ranges, which supports embodied tracking evaluation under both short-range maneuvers and longer-horizon target following.

In addition to statistical analysis, Fig.~\ref{fig:target_trajectory_simulator} provides qualitative trajectory examples to illustrate the typical geometry and waypoint density of target trajectories in the simulator.
The start waypoint is marked in green, the end waypoint is marked in red, intermediate waypoints are shown as black dots, and consecutive waypoints are connected by a yellow polyline.
These examples provide an intuitive visualization of the trajectory geometry and waypoint density in the benchmark.
\begin{figure*}[t]
  \centering
  \includegraphics[
    width=0.88\textwidth,
    trim=0 20 0 20,
    clip
  ]{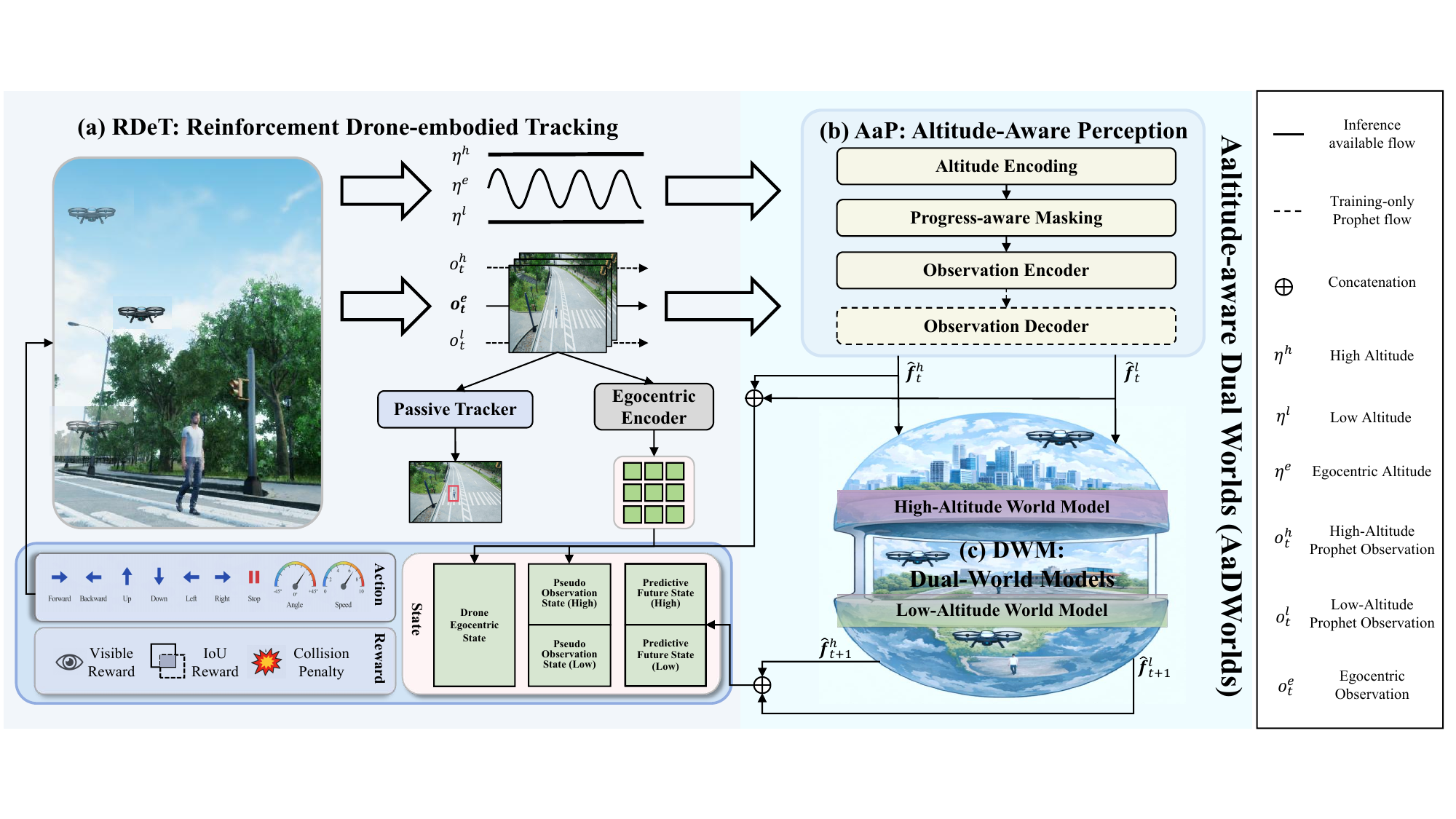}
  \vspace{-4pt}
  \caption{Overview pipeline of the proposed AaDWorlds framework for drone-embodied tracking. The framework consists of a reinforcement drone-embodied tracking (ReDeT) baseline, an altitude-aware perception (AaP) module, and dual world models (DWM). During inference, the pseudo high-altitude and low-altitude observation representations generated by AaP, together with the imagined future states predicted by DWM under both altitude regimes, complement the original drone-egocentric observation and provide auxiliary clues for current decision-making in closed-loop drone-embodied tracking.}
  \label{fig:framework}
  \vspace{-4pt}
\end{figure*}

\section{Method} 
\subsection{Overview}
As shown in Fig.~\ref{fig:framework}, the overall pipeline of our method consists of a reinforcement drone-embodied tracking (ReDeT) baseline and the altitude-aware dual worlds (AaDWorlds). Specifically, we formulate the drone-embodied tracking task into a reinforcement decision-making process (ReDeT). The ReDeT takes the egocentric drone observations to simultaneously predict the bounding-box of target via a conventional passive tracker and the flight action of drone via a reinforcement learning agent. Besides, the AaDWorlds consists of dual world models (DWM) and an altitude-aware perception (AaP) module. The DWM is equipped with a high-altitude world model and a low-altitude world model, which predict auxiliary imagined future observation features under both high- and low-altitude regimes. The AaP adaptively encodes, couples, and decodes altitude-aware representations at different altitudes. As a result, the imagined future states from DWM and pseudo egocentric observations from
AaP at both high and low altitudes complement original drone-egocentric observation and provide more comprehensive clues for current decision-making. The auxiliary clues at high and low altitudes simultaneously facilitate the view field, target detail, and obstacle avoidance, thus effectively alleviating the intrinsic altitude-mediated contradiction between visibility and safety.

To construct the dual world models (DWM), we always accompany the \textit{real} egocentric observation ($\bm{o}_{t}^{e}$) with a \textit{prophet} high-altitude observation ($o_{t}^{h}$) and a \textit{prophet} low-altitude observation ($o_{t}^{l}$) respectively at high and low altitudes ($\eta^{h}$ and $\eta^{l}$) observation during training stage, while we remove prophet observations during inference stage. As shown in Fig.~\ref{fig:framework}, the dashed line indicates training-only observations/structure and will be padded/removed during inference. Therefore, the (AaP) module is proposed to adaptively encode and couple altitude-aware representations at different altitudes, which will be used to decode \textit{pseudo} low-altitude and high-altitude representations ($\hat{\bm{f}}_{t}^{h}$ and $\hat{\bm{f}}_{t}^{l}$) during inference. In the following, we will introduce the details of ReDeT, DWM and AaP.

\subsection{Reinforcement Drone-embodied Tracking (ReDeT)}
In this paper, we formulate the drone-embodied tracking into a technical paradigm of decision-making process and construct a reinforcement baseline that combines reinforcement learning with passive tracking, termed reinforcement drone-embodied tracking (ReDeT). As shown in Fig.~\ref{fig:framework}a, the passive tracker tracks the target in current drone egocentric observations $o_{t}$. Simultaneously, the drone agent takes the latest $K$ drone egocentric observations $o_{t-K+1:t}$ and encodes them into the drone egocentric state ($s^e_t$). It predicts the next flight action ($a_t$) according to the current state ($s_t$). The action space includes 7 discrete movement commands (forward, backward, upward, downward, leftward, rightward, and stop) and two continuous adjustment commands for yaw-angle and speed, respectively. Thereafter, the drone agent moves in interactive 3D scene according to the flight action ($a_t$), and the spatial position and egocentric observation of drone agent refreshes accordingly.

The total reward of the ReDeT consists of 3 parts in consideration of bounding-box accuracy, target visibility, and collision punishment, \ie, $r =  r_{\mathrm{iou}} + \lambda_{\mathrm{vis}} r_{\mathrm{vis}} + \lambda_{\mathrm{col}} r_{\mathrm{col}}$. Specifically, $r_{\mathrm{iou}}=\mathrm{IoU}_t$,  which is the IoU between the predicted and ground-truth bounding-boxes. $r_{\mathrm{vis}}=v_t$, which is the visibility status of the target in current drone egocentric observation. $r_{\mathrm{col}}= -\mathbb{I}(\Delta c_t=1)$, which reflects the punishment for unwished collision, where  $c_t\in\{0,1\}$ indicates collision status and $\Delta c_t\ = 1$ indicates a collision-rising event. \(\lambda_{\mathrm{vis}}\) and \(\lambda_{\mathrm{col}}\) are degree controlling hyperparameters. Eventually, we adopt the above ReDeT as baseline framework for the proposed drone-embodied tracking task, and enhance it with auxiliary pseudo observation state ($\hat{\bm{f}}_{t}^{h}$ and $\hat{\bm{f}}_{t}^{l}$) and predictive future state ($\hat{\bm{f}}_{t+1}^{h}$ and $\hat{\bm{f}}_{t+1}^{l}$) during inference.

\subsection{Dual World Models (DWM)}
\begin{figure*}[!t]
    \centering
    \includegraphics[
        width=0.8\linewidth,
        trim=0 0 0 0,
        clip
    ]{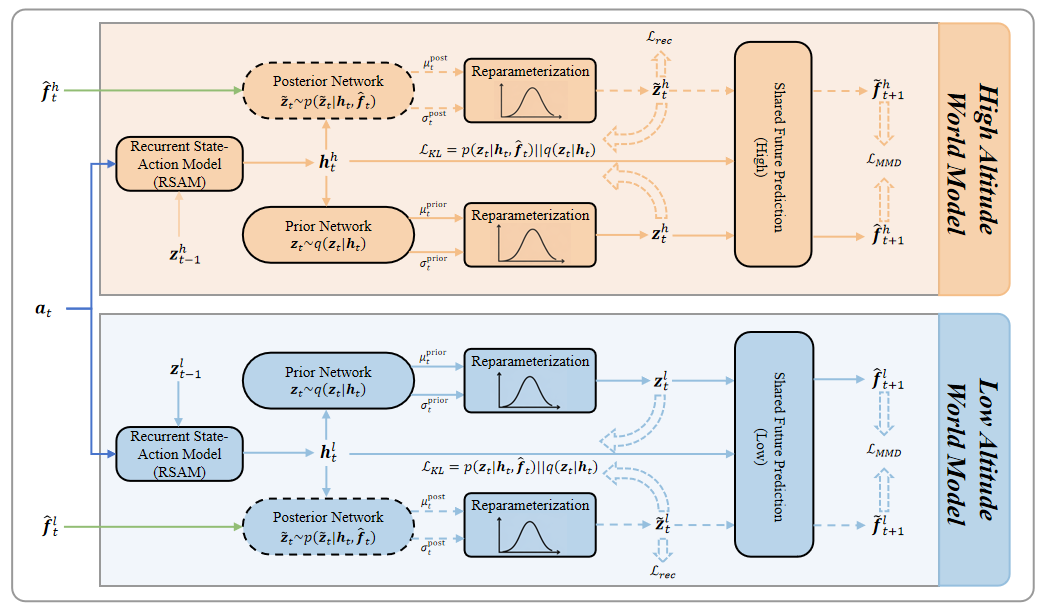}
    \vspace{-6pt}
    \caption{Detailed structure of the proposed Dual World Models (DWM). DWM consists of a High-Altitude World Model and a Low-Altitude World Model, each of which uses a recurrent state-action model (RSAM), posterior/prior networks, reparameterization, and a shared future prediction head to imagine altitude-specific future states for closed-loop drone-embodied tracking.}
    \label{fig:dwm}
    \vspace{-8pt}
\end{figure*}

To alleviate intrinsic altitude-mediated contradiction of visibility and safety in the drone-embodied tracking task, we build dual world models (DWM) to respectively simulate low-altitude and high-altitude evolution dynamics (Fig.~\ref{fig:dwm}). During inference, they are utilized to simultaneously imagine future states \((\hat{\bm{f}}^{h}_{t+1}, \hat{\bm{f}}^{l}_{t+1})\) at both low and high flight altitudes serving as auxiliary state clues.

Specifically, the DWM consists of two altitude-specific world models (\(\mathcal{W}^{h}\) and \(\mathcal{W}^{l}\)) for the high- and low-altitude evolution dynamics, respectively. For each altitude regime \(j\in\{h,l\}\), the world model $\mathcal{W}^{j}$ consists of a recurrent state-action model (RSAM), a posterior network, a prior network, and a future prediction head. Specifically, each model $\mathcal{W}^{j}$  maintains a recurrent hidden state \(\bm{z}_{t}^{j}\) and recurrent output state \(\bm{h}_{t}^{j}\) to summarize the historical state-action dynamics for world modeling and future prediction. Given the current action \(\bm{a}_{t}\) and the observation feature \(\hat{\bm{f}}_{t}^{j}\) at altitude regime $j$, the posterior network evolves via both the observation feature \(\hat{\bm{f}}_{t}^{j}\) and recurrent output state \(\bm{h}_{t}^{j}\), while the prior network evolves purely via recurrent output state \(\bm{h}_{t}^{j}\), \ie, 
\begin{equation}
\small
\begin{aligned}
\bm{h}_{t}^{j} 
&= \mathcal{G}^{j}(\bm{z}_{t-1}^{j}, \bm{a}_{t}), \\
\tilde{\bm{z}}_{t}^{j} 
&\sim p^{j}(\tilde{\bm{z}}_{t}^{j}\mid \bm{h}_{t}^{j}, \hat{\bm{f}}_{t}^{j}), \quad
\bm{z}_{t}^{j} \sim q^{j}(\bm{z}_{t}^{j}\mid \bm{h}_{t}^{j}).
\end{aligned}
\end{equation}
where \(\mathcal{G}^{j}(\cdot)\) denotes the recurrent state-action transition function. For simplicity, we assume the recurrent hidden state is in a form of Gaussian distribution. The posterior network (or prior network) predicts the mean and variance ($\mu, \sigma$) of its recurrent hidden state, and then generates a state from its distribution with the reparameterization trick \cite{kingma2014auto}. Finally, the shared future prediction head (\(\mathcal{D}^{j}(\cdot)\)) imagines corresponding future state feature \((\tilde{\bm{f}}_{t+1}^{j}, \hat{\bm{f}}_{t+1}^{j})\) the recurrent hidden state and recurrent output state, \ie, 
\begin{equation}
\small
\begin{aligned}
\tilde{\bm{f}}_{t+1}^{j} &= \mathcal{D}^{j}(\bm{h}_{t}^{j}, \tilde{\bm{z}}_{t}^{j}), \quad
\hat{\bm{f}}_{t+1}^{j} 
= \mathcal{D}^{j}(\bm{h}_{t}^{j}, \bm{z}_{t}^{j}).
\end{aligned}
\end{equation}
During training stage, the posterior network is trained to model evolution dynamics at altitude regime $j$ with external action-observation pairs, which is optimized by approaching prophet observation future \(\bm{f}_{t+1}^{j}\). The prior network is trained to mimic the evolution dynamics in the posterior network whose input is from external real-world action-observation pairs. During inference stage, the posterior network branch is removed and the prior network will only evolve with the external action input and historical hidden state without external observation feature anymore. The optimization objective for altitude regime $j$ is as follows:
\begin{equation}
\small
\begin{aligned}
\mathcal{L}^{j}_{\mathrm{DWM}}
=&~
\mathrm{KL}\!\left(
p^{j}(\tilde{\bm{z}}_{t}^{j}\mid \bm{h}_{t}^{j}, \hat{\bm{f}}_{t}^{j})
\,\|\, 
q^{j}(\bm{z}_{t}^{j}\mid \bm{h}_{t}^{j})
\right)
\\
&+
\lambda_{\mathrm{mmd}}
\mathrm{MMD}\!\left(
\tilde{\bm{f}}_{t+1}^{j},
\hat{\bm{f}}_{t+1}^{j}
\right) 
+
\lambda_{\mathrm{rec}}
\left\|
\tilde{\bm{z}}_{t}^{j}
-
\bm{z}_{t}^{j}
\right\|_{2}^{2}.
\end{aligned}
\end{equation}
where the KL and MMD respectively denote Kullback--Leibler divergence and maximum mean discrepancy, which respectively decrease the latent distribution distance and predictive representation distance between posterior and prior networks. The latent reconstruction loss is an $\ell_2$ loss between $\tilde{\bm{z}}_{t}^{j}$ and \(\bm{z}_{t}^{j}\), which governs the evolution dynamics learning. Eventually, the overall optimization objective for the DWM is \(\mathcal{L}_{\mathrm{DWM}}=\mathcal{L}_{\mathrm{DWM}}^{h}+\mathcal{L}_{\mathrm{DWM}}^{l}\). 
\begin{figure*}[h]
    \centering
    \includegraphics[
        width=0.8\linewidth,
        trim=0 0 0 0,
        clip
    ]{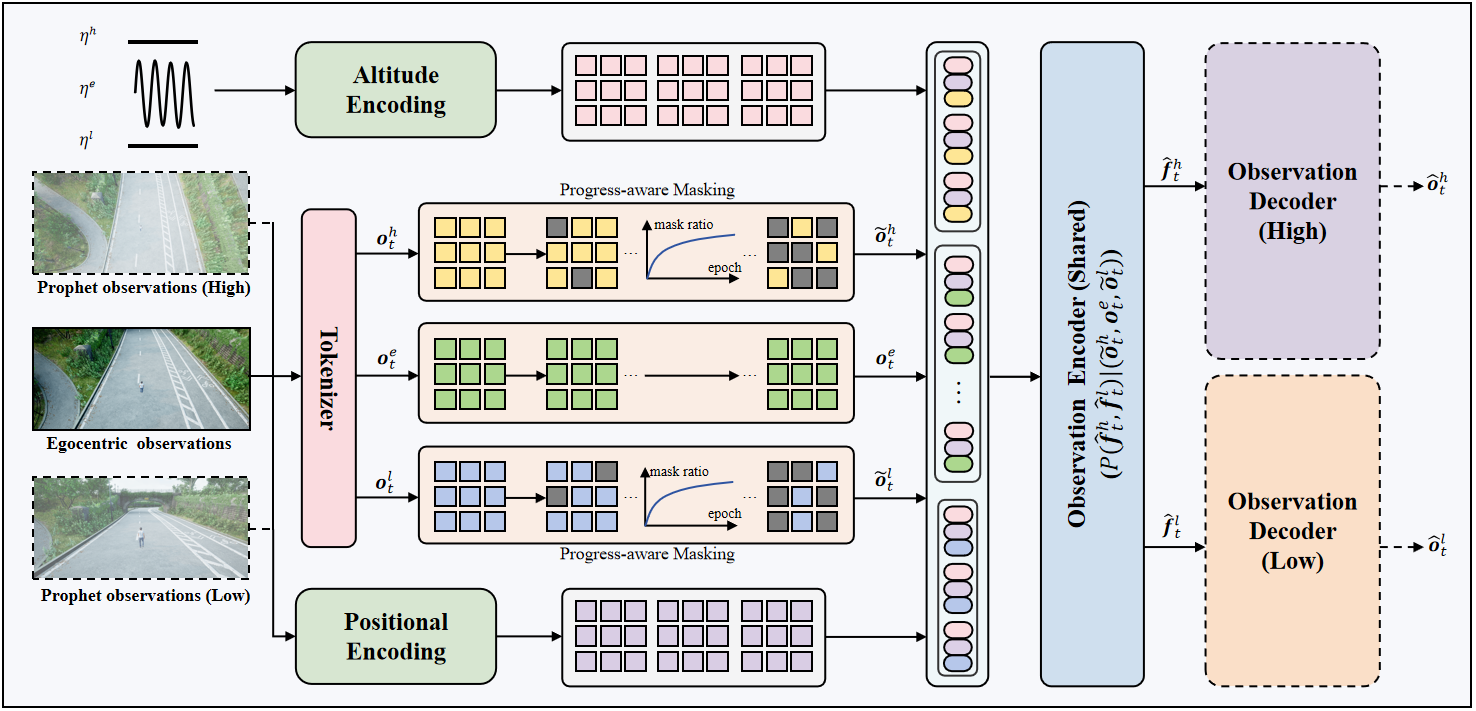}
    \vspace{-6pt}
    \caption{Detailed structure of the proposed Altitude-aware Perception (AaP) module. AaP takes the real egocentric observation together with prophet high-altitude and low-altitude observations during training, and exploits altitude encoding, tokenization, progress-aware masking, and a shared observation encoder to learn pseudo high-altitude and low-altitude representations for closed-loop drone-embodied tracking. The two observation decoders are used only during training to reconstruct altitude-specific observations and provide supervision.}
    \label{fig:aap}
    \vspace{-10pt}
\end{figure*}
\subsection{Altitude-aware Perception (AaP)}
\label{sec:aap}
The dual world models (DWM) require \textit{prophet} observations at high-altitude and low-altitudes (see Fig.~\ref{fig:framework} and Fig.~\ref{fig:dwm}) to model evolution dynamics during training, but the drone agent can only access a single real egocentric observation and these additional observations are unavailable during inference. To bridge this gap and adapt the inference workflow of DWM, we propose an altitude-aware perception (AaP) module in Fig.~\ref{fig:aap} adaptively encode and couple altitude-aware representations at different altitudes. It could be used to transform real egocentric observation into \textit{pseudo} low-altitude and high-altitude representations ($\hat{\bm{f}}_{t}^{h}$ and $\hat{\bm{f}}_{t}^{l}$) during inference.

Overall, the AaP is a form of masked autoencoders. During training, the prophet high-altitude observation \(o_{t}^{h}\) and low-altitude observation \(o_{t}^{l}\) first undergo a progress-aware masking (PaM) strategy while the real egocentric observation \(\bm{o}_{t}^{e}\) stays complete. The masking ratio (\(m(n)\)) in the PaM gradually increases from 0 to 1 as training proceeds (see Fig.~\ref{fig:aap} for details), so it just converges to the inference paradigm (no prophet observations) when training is finished. Then, they will be fed into a shared observation encoder for representation encoding and coupling, \ie
\begin{equation}
\small
\begin{aligned}
(\hat{\bm{f}}_{t}^{h},\hat{\bm{f}}_{t}^{l})
&\sim
P_{\theta}
\left(
\hat{\bm{f}}_{t}^{h},\hat{\bm{f}}_{t}^{l}
\mid
\bar{\bm{o}}_{t}^{h},
\bm{o}_{t}^{e},
\bar{\bm{o}}_{t}^{l}
\right), \\
\bar{\bm{o}}_{t}^{j}
&=
\mathcal{M}_{m(n)}
\left(
\bm{o}_{t}^{j}
\right),
\quad j\in\{h,l\}.
\end{aligned}
\end{equation}
where \(P_{\theta}(\cdot)\) denotes the shared observation encoder that couples the masked high-altitude prophet observation \(\bar{\bm{o}}_{t}^{h}\), the complete real egocentric observation \(\bm{o}_{t}^{e}\), and the masked low-altitude prophet observation \(\bar{\bm{o}}_{t}^{l}\). 
\(\mathcal{M}_{m(n)}(\cdot)\) is the progress-aware masking operation with masking ratio \(m(n)\). 
The outputs \(\hat{\bm{f}}_{t}^{h}\) and \(\hat{\bm{f}}_{t}^{l}\) denote pseudo high- and low-altitude current features used by the dual world models.
As a result, the AaP encodes and couples altitude-specific observations, and gradually learns to derive \textit{pseudo} high and low altitude representations from real egocentric observation. Finally, two training-only observation decoders respectively decode observations at the high and low altitudes \((\hat{\bm{o}}_{t}^{h}, \hat{\bm{o}}_{t}^{l})\)  forming altitude-specific supervisions, \ie

\begin{equation}
\small
\mathcal{L}_{\mathrm{AaP}}
=
\sum\nolimits_{j\in\{h,l\}}
\left\|
\hat{\bm{o}}_{t}^{j}
-
\bm{o}_{t}^{j}
\right\|_{2}^{2}.
\end{equation}
where \(\hat{\bm{o}}_{t}^{j}\) denotes the reconstructed observation decoded by the training-only observation decoder for altitude regime \(j\), and \(\bm{o}_{t}^{j}\) denotes the corresponding prophet observation. The objective encourages the reconstructed high- and low-altitude observations to approach their corresponding prophet observations. As a result, the AaP is capable of generating pseudo altitude-specific representations \((\hat{\bm{f}}_{t}^{h}, \hat{\bm{f}}_{t}^{l})\) during inference without any prophet observation for dual world models (DWM).

\section{Experiments}
\subsection{Implementation Details}
Regarding the main settings in our AaDWorlds framework, the backbone of the OSTrack tracking model is a Transformer-based visual encoder which is initialized with publicly available pretrained weights to ensure robust feature representation. Additionally, a pretrained ResNet-18 backbone is incorporated to further enhance the encoding capability of the policy head for real-time image streams. For the training configuration, we use the AdamW optimizer in the first two stages, with a base learning rate of 3e-4 and a weight decay of 0.05 for AaP pre-training, and a base learning rate of 1e-3 and a weight decay of 0.01 for DWM optimization. The AaP pre-training is conducted for 300 epochs with a streaming batch size of 8, while the DWM optimization involves 300 epochs on trajectory data collected under the DeTrack benchmark protocol. During the final ReDeT policy learning stage, the drone agent is trained for a total of 320,000 steps within the simulator using the PPO algorithm (learning rate 3e-4), with a clipping parameter of 0.2 and a generalized advantage estimation factor of 0.95. All methods are evaluated using visible rate (VR), mean IoU (mIoU), tracking rate (TR), and trajectory success rate (SR). Regarding the primary settings in our closed-loop drone-embodied tracking protocol. The embodied agent is equipped with a rigidly mounted RGB camera that renders images at a resolution of $640 \times 360$, featuring a $90^\circ$ horizontal field of view (FOV) and a fixed $0^\circ$ pitch angle. For all learning-based modules, input frames are resized to $256 \times 256$ to balance computational efficiency and perception granularity. In our altitude-aware setting, the high-altitude and low-altitude prophet observations are rendered at fixed flight heights of \(20\) m and \(3\) m, respectively. The simulator advances with a discrete time step of $\Delta t = 0.5$ s. During drone flight, the drone is subject to fixed control constraints in various scene styles, including a maximum linear speed of $8.0$m/s and a yaw-rate limit of $\pm 45^\circ$/s.

\subsection{Experimental Results}

\textbf{Baseline and comparing methods}. To establish a benchmark for the embodied tracking task in aerial interactive environments, we conduct performance comparison experiments based on several configurations including standard passive trackers, rule-based policies, reinforcement learning (RL) based policies, and the proposed AaDWorlds framework. These configurations reflect a logical progression of tracking paradigms. We start with traditional passive tracking methods represented by MOSSE~\cite{bolme2010mosse} and OSTrack~\cite{ye2022ostrack} which focus solely on target localization within pre-recorded frames. We then introduce an active but random baseline where a passive tracker (OSTrack) is combined with a random movement strategy to demonstrate the impact of stochastic egocentric transitions on tracking stability. Beyond random exploration, we also incorporate a heuristic-based policy that utilizes a rule-based controller to adjust the drone position for target centering. Furthermore, we evaluate more sophisticated closed-loop systems that integrate specialist trackers with reinforcement learning policies (such as AC~\cite{konda2000actorcritic} and PPO~\cite{schulman2017ppo} architectures) to establish a competitive benchmark for drone-embodied tracking. Building upon the baseline (PPO + OSTrack), the proposed AaDWorlds framework further integrates altitude aware perception and dual world model strategies to handle complex 3D dynamic environments and provide predictive pseudo observations. Finally, we introduce a prophet policy as the upper bound method, which leverages prophet ground-truth trajectory information to maintain near-optimal positioning without any perception noise. This upper bound allows us to quantify the performance gap and evaluate how effectively our tracking policy approximates optimal tracking behavior in complex environments. We evaluate the performance of these methods on the DeTrack benchmark across all 4 primary evaluation metrics in the Section~\ref{metrics}.  

\begin{table}[!t]
\centering
\vspace{-5pt}
\captionsetup{font=footnotesize, skip=2pt}
\caption{Experimental results on the DeTrack (\%).}
\label{tab:overall_results}
\vspace{1pt}

\scriptsize
\setlength{\tabcolsep}{2.0pt}
\renewcommand{\arraystretch}{0.92}
\resizebox{\linewidth}{!}{
\begin{tabular}{@{}llcccc@{}}
\toprule
\textbf{Paradigm} & \textbf{Method} & \textbf{VR}$\uparrow$ & \textbf{mIoU}$\uparrow$ & \textbf{TR}$\uparrow$ & \textbf{SR}$\uparrow$ \\
\midrule
\multirow{2}{*}{Passive} 
& MOSSE~\cite{bolme2010mosse} & 9.32 & 7.17 & 6.31 & 1.41 \\
& OSTrack~\cite{ye2022ostrack} & 9.34 & 7.39 & 6.32 & 1.41 \\
\midrule
\multirow{2}{*}{Rule-based}  
& Random + OSTrack & 10.35 & 8.32 & 8.42 & 3.42 \\
& Heuristic + OSTrack & 12.32 & 10.25 & 11.21 & 7.42 \\
\midrule
\multirow{3}{*}{RL-based} 
& AC + OSTrack & 24.58 & 14.34 & 18.64 & 12.97 \\
& PPO + OSTrack (Baseline) & 25.88 & 15.67 & 19.07 & 14.08 \\
& \textbf{AaDWorlds (Ours)} & \textbf{29.00} & \textbf{17.06} & \textbf{22.57} & \textbf{17.66} \\
\midrule
Upper-bound 
& Prophet + OSTrack & 96.28 & 77.28 & 87.06 & 95.73 \\
\bottomrule
\end{tabular}
}

\vspace{-8pt}
\end{table}

\textbf{Performance comparison on the DeTrack benchmark}. 
(1) \textit{Performance of passive trackers degrades significantly on the DeTrack benchmark}: Although conventional passive  trackers (such as MOSSE~\cite{bolme2010mosse} and OSTrack~\cite{ye2022ostrack}) exhibit great performance on the pre-recorded tracking datasets (e.g., OSTrack reports 70.7\% AUC on UAV123~\cite{mueller2016uav123} and 83.9\% AUC on TrackingNet~\cite{muller2018trackingnet}), they achieve very poor tracking performance on our DeTrack benchmark tailored for drone-embodied tracking task (Table~\ref{tab:overall_results}). The performance degradation is primarily attributed to the intrinsic characteristics of the DeTrack benchmark. These challenges collectively exacerbate the difficulty of target localization and flight control and eventually constitute the fundamental reasons for the observed performance degradation. (2) \textit{Rule-based policy brings some but limited improvement}: With the aid of the heuristically rule-based closed-loop drone controlling policy, the performance improves by some degree over passive tracker OSTrack~\cite{ye2022ostrack}, indicating the importance of closed-loop drone controlling during the drone-embodied tracking. (3) \textit{RL-based policies bring promising improvement and our AaDWorlds achieves the best performance}. The reinforcement learning policies (AC~\cite{konda2000actorcritic} and PPO~\cite{schulman2017ppo}) achieves much promising improvement for passive trackers and the PPO policy performs better than the AC policy, where PPO+OSTrack is chosen as the baseline of our AaDWorlds framework. Eventually, our AaDWorlds achieves best performance across all evaluation metrics that demonstrates the stronger adaptability during drone-embodied tracking. In detail, it improves the VR from 25.88\% to 29.00\%, the mIoU from 15.67\%  to 17.06\% , the TR from 19.07\%  to 22.57\%, and the SR from 14.08\%  to 17.66\%. The performance gains attribute to our altitude aware perception module and dual world model strategy. The dual world models predict imagined future features under both high- and low-altitude regimes to complement drone-egocentric observation, which simultaneously facilitates the view field, target details, and obstacle avoidance during drone-embodied tracking.

\subsection{Ablation Studies}
To analyze the effectiveness of each primary component in our AaDWorlds framework, we conduct ablation studies on the DeTrack benchmark. Starting from the PPO + OSTrack baseline, we progressively introduce the Altitude-aware Perception (AaP) module and the Dual World Models (DWM) under different combinations. As shown in Table~\ref{tab:ablation_main}, each module brings performance improvements across all metrics. The AaP improves the representation of altitude-aware current features under extensive variation, while the DWM further enhances decision-making by providing imagined future features for both high- and low-altitude regimes. When combining the AaP and DWM together, the model achieves the best performance, demonstrating the effectiveness of our AaDWorlds framework.

\begin{table}[tbp]
\centering
\caption{Ablation studies on the DeTrack.}
\label{tab:ablation_main}

\begin{tabular}{@{}ccccccc@{}}
\toprule
\textbf{Base} & \textbf{AaP} & \textbf{DWM} & \textbf{VR}$\uparrow$ & \textbf{mIoU}$\uparrow$ & \textbf{TR}$\uparrow$ & \textbf{SR}$\uparrow$ \\
\midrule
$\checkmark$ & $\times$ & $\times$ & 25.88\% & 15.67\% & 19.07\% & 14.08\% \\
$\checkmark$ & $\checkmark$ & $\times$ & 27.79\% & 15.78\% & 19.42\% & 14.13\% \\
$\checkmark$ & $\times$ & $\checkmark$ & 27.64\% & 16.24\% & 20.15\% & 14.61\% \\
$\checkmark$ & $\checkmark$ & $\checkmark$ & \textbf{29.00\%} & \textbf{17.06\%} & \textbf{22.57\%} & \textbf{17.66\%} \\
\bottomrule
\end{tabular}
\end{table}

\subsection{Influence analysis of the altitude combinations}
\label{sec:altitude_pair_analysis}

To further analyze the effect of flight altitude on altitude-aware prophet observations, we conduct an altitude-pair sensitivity analysis by varying the low- and high-altitude settings used in AaP and DWM. In the altitude-aware setting, the low-altitude prophet observation usually provides larger target scale and richer local details, but overly low viewpoints may introduce stronger near-view occlusion and unstable perspective changes. In contrast, the high-altitude prophet observation provides a wider field of view and more global motion context, while overly high viewpoints may make the target too small and visually ambiguous. As shown in Table~\ref{tab:altitude_pair_ablation}, the \(3\)m/\(20\)m pair achieves the best performance across all evaluation metrics. Compared with \(1\)m/\(20\)m, it avoids excessive near-view occlusion and perspective instability; compared with \(6\)m/\(20\)m, it better preserves the target-detail advantage of the low-altitude branch. For the high-altitude branch, \(10\)m provides relatively limited global context, whereas \(40\)m weakens fine-grained tracking cues due to the smaller target scale. The lower performance of the corner combinations further indicates that the two altitude branches should provide complementary local and global information rather than overly close or overly distant observations. These results suggest that the \(3\)m/\(20\)m setting forms a more suitable trade-off for altitude-aware perception and future imagination in drone-embodied tracking.

\begin{table}[tbp]
\centering
\caption{Influence Analysis of altitude combinations in the proposed AaDWorlds on the DeTrack benchmark.}
\label{tab:altitude_pair_ablation}

\begin{tabular}{@{}cc cccc@{}}
\toprule
\textbf{Low Alt.} & \textbf{High Alt.} 
& \textbf{VR}$\uparrow$ 
& \textbf{mIoU}$\uparrow$ 
& \textbf{TR}$\uparrow$ 
& \textbf{SR}$\uparrow$ \\
\midrule
\(1\,\mathrm{m}\) & \(10\,\mathrm{m}\) & 25.62\% & 14.91\% & 19.34\% & 13.22\% \\
\(1\,\mathrm{m}\) & \(20\,\mathrm{m}\) & 27.52\% & 16.28\% & 21.12\% & 16.01\% \\
\(1\,\mathrm{m}\) & \(40\,\mathrm{m}\) & 24.96\% & 14.58\% & 18.83\% & 12.57\% \\
\midrule
\(3\,\mathrm{m}\) & \(10\,\mathrm{m}\) & 27.71\% & 16.17\% & 21.08\% & 15.92\% \\
\textbf{\(3\,\mathrm{m}\)} & \textbf{\(20\,\mathrm{m}\)} & \textbf{29.00\%} & \textbf{17.06\%} & \textbf{22.57\%} & \textbf{17.66\%} \\
\(3\,\mathrm{m}\) & \(40\,\mathrm{m}\) & 26.93\% & 15.82\% & 20.61\% & 15.34\% \\
\midrule
\(6\,\mathrm{m}\) & \(10\,\mathrm{m}\) & 26.21\% & 15.25\% & 19.76\% & 13.68\% \\
\(6\,\mathrm{m}\) & \(20\,\mathrm{m}\) & 27.86\% & 16.39\% & 21.34\% & 16.24\% \\
\(6\,\mathrm{m}\) & \(40\,\mathrm{m}\) & 25.34\% & 14.79\% & 19.07\% & 12.95\% \\
\bottomrule
\end{tabular}
\end{table}

\subsection{Qualitative Visualization}
\begin{figure*}[!t]
    \centering
    \includegraphics[
        width=0.98\textwidth,
        trim=20 50 20 50,
        clip
    ]{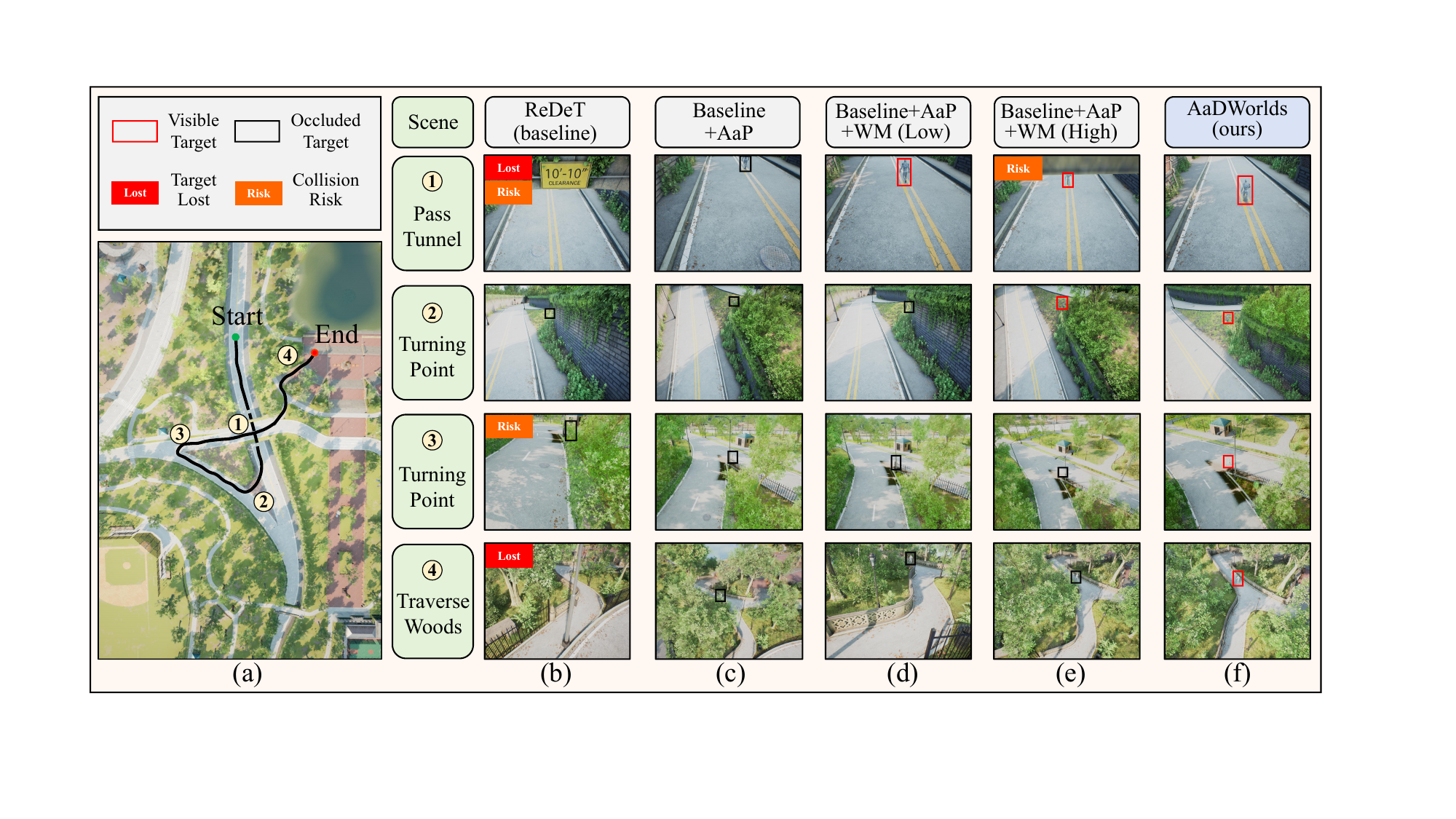}
    \vspace{-4pt}
    \caption{Qualitative visualization of target visibility and flight safety under four representative scenes. (a) The  overview of target trajectory. 
    (b)-(f) The results comparison between our AaDWorlds and 4 methods variants. 
    }
    \label{fig:qualitative_visualization}
    \vspace{-8pt}
\end{figure*}

As shown in Fig.~\ref{fig:qualitative_visualization}(b), the ReDeT baseline performs worse since the current egocentric observation is insufficient for reliable closed-loop tracking. For example, in the Pass Tunnel and Traverse Woods scenes, the  ReDeT baseline encounters target loss and unsafe close-range flight around tunnel structures and dense vegetation (see Fig.~\ref{fig:qualitative_visualization}(b-1) and Fig.~\ref{fig:qualitative_visualization}(b-4)). In the two Turning Point scenes, the drone adjusts its viewpoint only after the target direction has changed, leading to delayed direction following and unstable target observability (see Fig.~\ref{fig:qualitative_visualization}(b-2) and Fig.~\ref{fig:qualitative_visualization}(b-3)). After introducing the proposed AaP module, the variant Baseline + AaP in Fig.~\ref{fig:qualitative_visualization}(c) improves the perception of target scale and local scene structure through altitude-aware current features. Compared with the results from ReDeT baseline in Fig.~\ref{fig:qualitative_visualization}(b), the target can be kept more observable in several scenes. However, since this variant still lacks future dynamics, it remains difficult to anticipate turning changes and near-obstacle interactions before the viewpoint becomes degraded. Furthermore, the proposed world models at low and high altitudes effectively alleviate this issue and lead to different scene-dependent behaviors. On the one hand, aiding with the low-altitude world model (Baseline+AaP+WM (Low)), Fig.~\ref{fig:qualitative_visualization}(d-1) and Fig.~\ref{fig:qualitative_visualization}(d-4) preserve more target-detail clues in the Pass Tunnel and Traverse Woods scenes, but the drone in this variant tends to stay closer to tunnel structures, road boundaries, and vegetation, leaving limited safety margins. On the other hand, aiding with the high-altitude world model (Baseline+AaP+WM (High)), Fig.~\ref{fig:qualitative_visualization}(e-2) and Fig.~\ref{fig:qualitative_visualization}(e-3) show better responses in the Turning Point scenes, because the wide-view imagined states at high altitude provide more global motion context for earlier direction adjustment, but this variant weakens local target-detail and obstacle-structure cues, and thus still suffers from risky behavior in narrow or cluttered areas, such as the Pass Tunnel scene in Fig.~\ref{fig:qualitative_visualization}(e-1).  Eventually, our AaDWorlds in Fig.~\ref{fig:qualitative_visualization}(f) combines world models from both low and high altitudes, allowing the agent to use both future target-detail clues in the low altitude and future wide-view context in the high altitude. Compared with the variant in Fig.~\ref{fig:qualitative_visualization}(d) and Fig.~\ref{fig:qualitative_visualization}(e), our AaDWorlds keeps the target observable and simultaneously maintains safer distances from surrounding structures across all 4 scenes, effectively alleviating the intrinsic altitude-mediated contradiction between target visibility and flight safety during drone-embodied tracking.

\section{Conclusion} 
In this paper, we introduce a benchmark (DeTrack) for closed-loop drone-embodied tracking in interactive 3D environments, and propose an altitude-aware dual-world (AaDWorlds) framework for this task. The DeTrack differs from conventional passive aerial tracking by considering perception-action coupling, mixed occlusions, altitude-dependent visibility-safety trade-offs, and joint tracking-obstacle avoidance. To address the intrinsic altitude-mediated contradiction, the AaDWorlds combines dual world models with an altitude-aware perception module, providing altitude-aware current features and imagined future features for action prediction. Experimental results on the DeTrack benchmark show that our AaDWorlds framework significantly improves the performance across all metrics.

\bibliographystyle{IEEEtran}
\bibliography{references}

\end{document}